%% file: main_cameraready.tex
\pdfoutput=1

\documentclass[11pt]{article}

\usepackage[final]{acl}

\usepackage{times}
\usepackage{latexsym}

\usepackage[T1]{fontenc}

\usepackage[utf8]{inputenc}

\usepackage{microtype}

\usepackage{inconsolata}

\usepackage{graphicx}

\usepackage{booktabs}
\usepackage{multirow}
\usepackage{amsmath}
\usepackage{amssymb}
\usepackage{subcaption}
\usepackage{enumitem}
\usepackage{graphicx}
\usepackage{todonotes}
\usepackage{subcaption}

\usepackage{tikz}
\usepackage{pgfplots}
\usepgfplotslibrary{groupplots}
\pgfplotsset{compat=1.17}

\title{Prepending or Cross-Attention for Speech-to-Text? \\ An Empirical Comparison}
% ...

\author{Tsz Kin Lam$^{1\dagger}$ , Marco Gaido$^{2\dagger}$ , Sara Papi$^{2}$ , Luisa Bentivogli$^{2}$ , Barry Haddow$^{1}$
  \\$^{1}$School of Informatics, University of Edinburgh \\ 
  $^{2}$Fondazione Bruno Kessler \\
  \texttt{\normalsize\{tlam, bhaddow\}@ed.ac.uk } \\
  \texttt{\normalsize\{mgaido, spapi, bentivo\}@fbk.eu} \\
 }

\begin{document}
\maketitle
\begin{abstract}
Following the remarkable success of Large Language Models (LLMs) in NLP tasks, there is increasing interest in extending their capabilities to speech---the most common form of communication. The most widespread approach to integrating speech into LLMs is dense feature prepending (DFP), which prepends the projected speech representations to the textual representations, allowing end-to-end training with a speech encoder. This raises questions about the need for a sophisticated speech encoder for DFP and how its performance compares with a standard encoder-decoder (i.e., cross-attention) architecture. We compare DFP and cross-attention under a variety of configurations, such as CTC compression, sequence-level knowledge distillation, on monolingual, bilingual, and multilingual models. To perform a controlled architectural comparison, we train all models from scratch rather than using large pretrained models and use comparable data and parameter settings, testing speech-to-text recognition (ASR) and translation (ST) on MuST-C v1.0 and CoVoST2 datasets. Despite the wide adoption of DFP, our results do not indicate a clear advantage of DFP over cross-attention.
\end{abstract}

\section{Introduction}
\def\thefootnote{$\dagger$}\footnotetext{These authors contributed equally to this work}\def\thefootnote{\arabic{footnote}}
As the NLP community has witnessed the emergence of Large Language Models (LLMs) and their remarkable performance in tackling NLP tasks \citep{radford2019language,touvron2023llama,jiang2023mistral,team2024gemma}, there is increasing interest in extending their capabilities to other modalities, such as audio \citep{latif2023sparks,chu2023qwen,huang2024audiogpt} and images \citep{radford2021learning,team2024chameleon}, to broaden their applicability. One of the most natural extensions of LLMs is to inject them with speech -- the most common form through which humans express their language \citep{munteanu-etal-2013-hci} -- to exploit the LLM's linguistic fluency and skills to tackle speech-to-text (S2T) tasks, such as automatic speech recognition (ASR) and S2T translation (ST).

This goal has been predominantly pursued by dense feature prepending (DFP) which adapts the embedded speech representations -- obtained using the encoder of a Speech Foundation Model (SFM) or a speech encoder trained from scratch -- to the input feature space of an LLM via a modality adapter and, optionally, a length adapter \citep{10389705,pan2023cosmic,wang2023slm} and prepends them to a textual prompt describing the tasks to be performed \citep{gaido-etal-2024-speech}. Existing works on DFP mainly train the speech encoder, the modality adapter, and low-rank adapters in the LLM in an end-to-end fashion \cite{chen-etal-2024-llast,hu2024wavllm}. Throughout the paper, we refer to this DFP solution leveraging a speech encoder or SFM as \emph{decoder-prepend}.

The effectiveness of \emph{decoder-prepend} has recently been questioned on the basis of the analogy with classic S2T encoder-decoder models \citep{chen2024bestowefficientstreamablespeech,zelasko2024emmett}, where the integration of the encoder output into the decoder is performed through \emph{cross-attention} modules \citep{Ao2021SpeechT5,radford2023robust,barrault2023seamlessm4t}. In addition, the outstanding performance of decoder-only LLMs on NLP tasks traditionally handled by encoder-decoder models has motivated the exploration of \emph{decoder-only} S2T models \citep{10389705,gupta24_interspeech}, which can be regarded as DFP solutions that question the need for a speech encoder and directly prepend speech features to the text embeddings. In this context, \citet{10389705,gupta24_interspeech} highlighted the crucial role of relaxing the causal\footnote{The \textit{causality} property refers to prohibiting a token from accessing successive tokens in the sequence, both at training and inference time, and is typically achieved through a diagonal masking matrix in the self-attention computation of Transformer decoders \citep{vaswani2017attention}.} masking in the self-attention modules typical of autoregressive models for the speech features, allowing them to look at each other freely, rather than being restricted to only previous elements. Notably, \citet{gupta24_interspeech} claimed that this approach enables \emph{decoder-only} models to even surpass encoder-decoder ones on the ASR task. On the contrary, to the best of our knowledge, no investigation on the effect of the causality property has been carried out for \textit{decoder-prepend} models.

With the goal of shedding light on the strengths and weaknesses of DFP solutions, we \textit{i)} systematically compare the two DFP-based architectures (\emph{decoder-prepend} and \emph{decoder-only}) with the standard encoder-decoder architecture using cross-attention, not only in terms of performance but also computational demands; \textit{ii)} perform this comparison under a variety of relevant configurations, covering monolingual, bilingual and multilingual settings and including widely adopted techniques such as
speech sequence length reduction (using the CTC compression mechanism -- \citet{liu2020bridging,gaido-etal-2021-ctc}) and sequence-level knowledge distillation or seqKD \cite{kim-rush-2016-sequence};
and \textit{iii)} conduct an in-depth study of the causality properties of DFP architectures.

To ensure a sound and fair comparison across architectures, we train them \textit{from scratch} on the same two datasets: MuST-C v1.0 \cite{di-gangi-etal-2019-must} and CoVoST2 \cite{wang2021covost}.
This choice of not using large-scale pretrained models offers a number of advantages. First, it prevents our results from being influenced by \textit{i)} the specific features and capabilities of pretrained models \citep{verdini2024connectspeechfoundationmodels}, and \textit{ii)} the lack of a well-established method for integrating speech encoder with LLMs via cross-attention, an area still in its early research stages \citep{chen2024bestowefficientstreamablespeech,zelasko2024emmett}. 
Furthermore, using small models instead of large-scale ones allows us to investigate the impact of all the aforementioned configurations within a reasonable computational budget.

Our experiments, carried out on two S2T tasks -- ASR and ST -- and covering a total of 12 language directions, demonstrate that:

\begin{itemize}
    \item Cross-attention and decoder-prepend lead to overall similar results in terms of quality on both ASR and ST, with the first being slightly more efficient in terms of generation speed and GPU memory footprint, both outperforming decoder-only models on all aspects.
    \item DFP benefits more than cross-attention from CTC compression in both ASR and ST.
    \item The inclusion of a speech encoder affects the causality behaviour of DFP models. Applying causal masking on the speech inputs hurts both the ASR and ST performances of decoder-only. In contrast, decoder-prepend slightly benefits from masking. 
\end{itemize}

While the scalability of the findings to large-scale models has to be confirmed (see Limitations), we believe that these findings can inform future research on integrating dense speech features into LLM. We release the code used in our experiments under the Apache 2.0 License at: \url{https://github.com/hlt-mt/FBK-fairseq/}.

\begin{figure*}
    \centering
    \begin{subfigure}[t]{0.29\textwidth}
            \centering
            \caption{Cross-Attention}
            \label{fig:encoder-decoder}
            \includegraphics[width=\linewidth,  keepaspectratio]{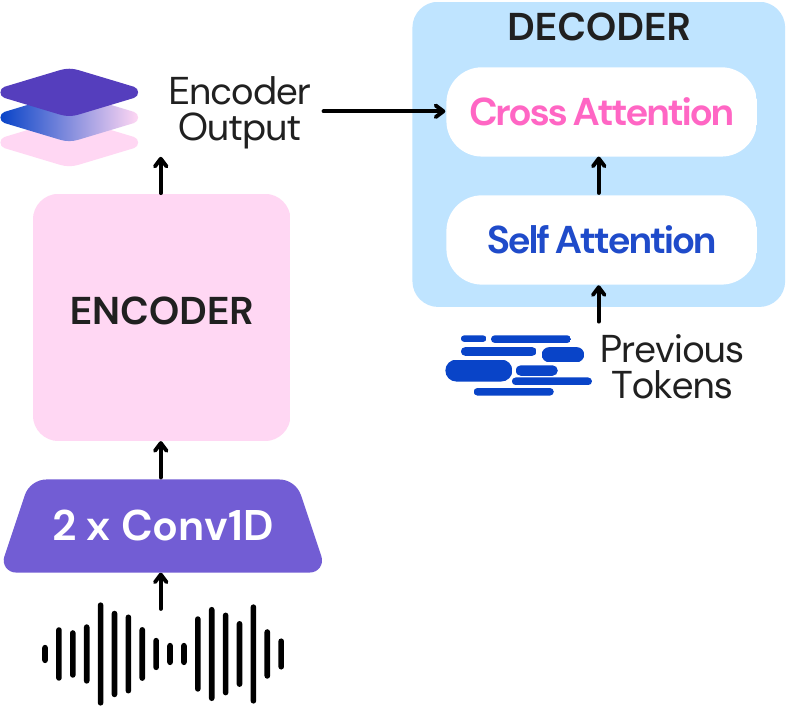}
        \end{subfigure}
        \hfill
        \begin{subfigure}[t]{0.3\textwidth}
            \centering
            \caption{Decoder-Only}
            \label{fig:decoder-only}
            \includegraphics[width=\linewidth, keepaspectratio]{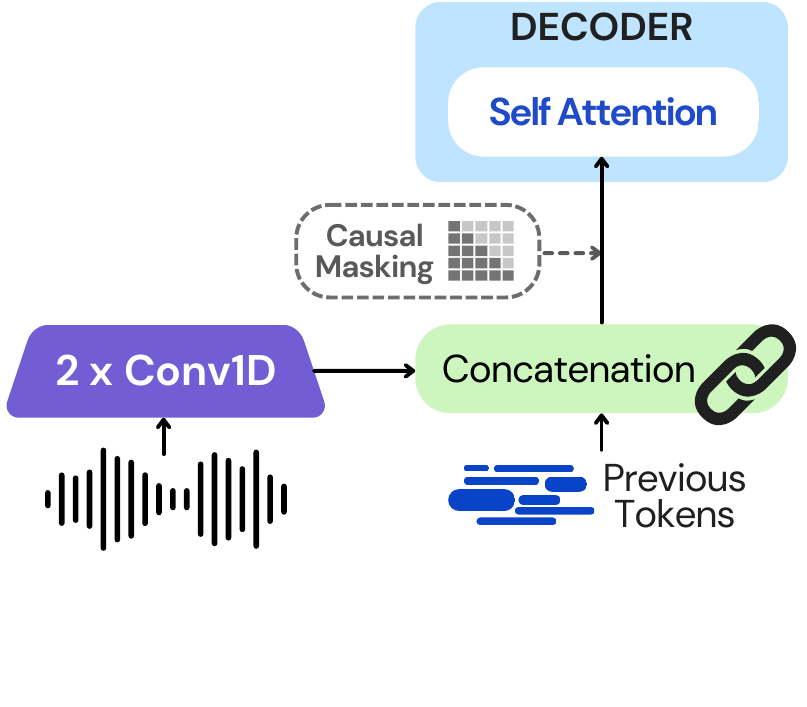}
        \end{subfigure}
        \hfill
        \begin{subfigure}[t]{0.35\textwidth}
            \centering
            \caption{Decoder-Prepend}
            \label{fig:decoder-prepend}
            \includegraphics[width=\linewidth,  keepaspectratio]{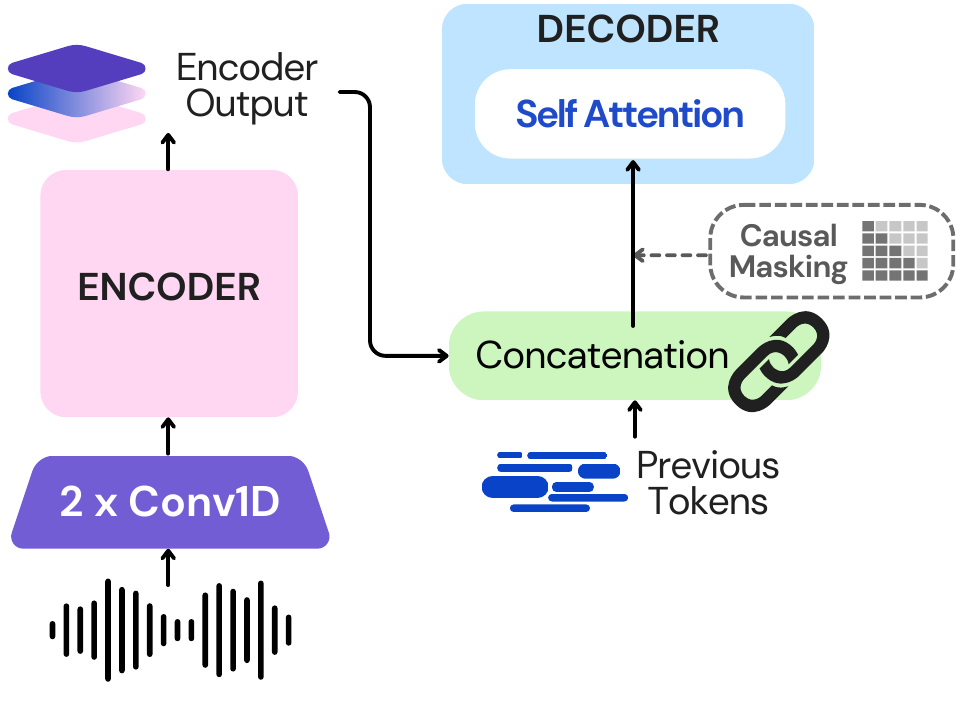}
        \end{subfigure}
    \caption{Representation of the architectures analyzed in the paper. Both (a) and (c) are based on encoder-decoder architecture but (a) uses cross-attention, whereas (c) uses DFP. Secondly, both (b) and (c) uses DFP, but (c) contains a speech encoder, making it not decoder-only. The (audio) causal masking can be applied to both the previous tokens and the audio sequence or only to the previous tokens.}
    \label{fig:architectures}
\end{figure*}

\section{Background}

\subsection{(Cross-)Attention-based encoder-decoder}
The cross-attention based encoder-decoder has been one of the major research directions for S2T \cite{chan2015listen,berard2016listen,weiss2017sequence,bansal-etal-2017-towards,fang-etal-2022-stemm,tsiamas-etal-2024-pushing}. In addition to its end-to-end (E2E) properties, such as a simpler pipeline over the traditional methods and E2E optimization, cross-attention allows full attention on the sequences, making it more attractive than CTC \cite{graves2006connectionist} and Transducer for learning sequences with switching word order, as in machine translation (MT) \cite{sperber-paulik-2020-speech,li2022recent}.


\subsection{DFP: modelling S2T with decoder-only language models}

With the tremendous success of decoder-only language models (LM) for modelling text, there have been explorations of using them for modelling S2T, such that the speech (source) embeddings are passed to the decoder via prepending to the target text embeddings rather than using cross-attention. In this work, we divide DFP methods into two categories: decoder-only and decoder prepend.

\paragraph{Decoder-only S2T.} We refer to decoder-only S2T as a model that has a length adapter (e.g., strided convolutions) for the speech inputs, but not a \textit{deep} speech encoder, before prepending.
These works include \citet{10389705}, which claims that decoder-only models can match the performance of encoder-decoder ones with fewer parameters on multilingual ST, and \citet{gupta24_interspeech}, which trains a large-scale ASR decoder-only model comparing it with openly-available models. In contrast, our experiments are not limited to a single task or setting, and we always compare models trained on the same data.

\paragraph{Decoder-prepend S2T.} When the projected speech embeddings are fed directly from a speech encoder or SFM, we refer to the model as decoder-prepend rather than decoder-only to emphasize its reliance on a speech encoder. There have been more works on this line, including ASR \cite{lakomkin2024end,hono-etal-2024-integrating,fathullah2024prompting,tsunoo2024decoder,tsunoo2023decoder}, ST \cite{huang2024investigating,chen-etal-2024-llast} and multi-tasks \cite{chen2024bestowefficientstreamablespeech,zelasko2024emmett,chen2024salm} systems. Despite their valuable insights,
their experiments are based on large pretrained models and lack transparency and comparability. In contrast, we remove such dependency, enabling a clear comparison with cross-attention models.

\section{Methods}

\subsection{Encoder-Decoder with Cross Attention}
Transformer-based architectures \citep{vaswani2017attention} are encoder-decoder sequence-to-sequence models, where the encoder maps the input sequence $\mathbf{X}=[x_1,...,x_n]$ into an internal representation or \textit{encoder output} (Figure \ref{fig:encoder-decoder}), which are then processed by the decoder to generate the output sequence $\mathbf{Y}=[y_1,...,y_m]$. Both encoder and decoder are composed of a stack of Transformer-based layers that exploit dot-product \textit{attention} ($A$) \citep{7472621} as the core mechanism, which is formulated as:
\begin{equation*}
    A(Q,K,V) = softmax \left( \frac{QK^T}{\sqrt{d_k}} \right) V 
\end{equation*}
where $Q$ is the query matrix, $K$ is the key matrix, $V$ is the value matrix, and $d_k$ is the dimension of $K$. In the encoder, the $Q$, $K$, and $V$ matrices are all obtained from the input sequence $\mathbf{X}$, and $A$ is called \textit{self-attention} ($A_s$)\footnote{Thus, we only display one input argument for readability.}. In the decoder, apart from the self-attention, there is another attention mechanism called \textit{cross-attention} ($A_c$) that links the encoder with the decoder representations. In this case, the $Q$ matrix of the $l_d$ layer is obtained from the output of the self-attention of the same layer, which takes as input the previous decoder layer output $H_{l_d-1}$, where $H_0$ is the sequence of embeddings of the previously generated output $\mathbf{Y}_{0, ..., i-1}$. The $K$ and $V$ matrices, instead, are taken from the encoder output $Enc(\mathbf{X})$. The Transformer decoder layer $D_{l_d}$ is completed by a feed-forward network ($FFN$) composed of two linear layers. As such, the output of the cross-attention-based encoder-decoder corresponds to the output of the last decoder layer $H_{L_d}$:
\begin{equation*}
\begin{split}
FFN(A_c(A_s(H_{L_{d}-1}), Enc(\mathbf{X}), Enc(\mathbf{X}))), \\
\text{where} \quad H_0 = \mathbf{Y}_{0, ..., i-1}
\end{split}
\end{equation*}
In the context of speech processing, the input sequence $\mathbf{X}$ is an audio segment downsampled
by a factor of 4 with two Convolutional layers before feeding it to the stack of encoder layers \citep{berard_2018,di-gangi-etal-2019-enhancing}. The downsampling 
maps the input representation into a shorter sequence suitable for processing, as audio is $\sim$10 times longer than the corresponding text sequence. 

\subsection{Decoder-only and Decoder-prepend}
In the decoder-only architecture \citep{brown2020language} for speech-to-text processing \citep{chen2023x}, the input audio sequence $\mathbf{X}$ is not processed by an encoder but its downsampled representation is directly fed into the decoder after being concatenated\footnote{We use prepending and concatenation interchangeably.} with the previously emitted tokens $\mathbf{Y}_{0, ..., i-1}$ (Figure \ref{fig:decoder-only}). The output of the decoder-only model $H_{L_d}$ can be expressed as:
\begin{equation*}
\begin{split}
FFN(A_s(H_{L_{d}-1})),
\text{where} \quad H_0 = \mathbf{X} \mathbin\Vert \mathbf{Y}_{0, ..., i-1}
\label{eq:dec-only}
\end{split}
\end{equation*}
\noindent in which $\mathbin\Vert$ is the concatenation operator. In this case, the cross-attention is dropped and self-attention is applied to both the input audio sequence $\mathbf{X}$ and the previous tokens $\mathbf{Y}_{0, ..., i-1}$.

The decoder-prepend (Figure \ref{fig:decoder-prepend}) architecture operates similarly to decoder-only, with the only difference that it exploits the representation obtained from a speech encoder instead of the raw speech features $\mathbf{X}$, as in the encoder-decoder models equipped with cross-attention. This corresponds to $H_0 = Enc(\mathbf{X}) \mathbin\Vert \mathbf{Y}_{0, ..., i-1}$ in the previous equation. A notable difference between decoder-prepend and decoder-only is that in decoder-prepend the audio frames can attend to each other and interact in the encoder before concatenation (prepending).

\subsection{Audio Causal Masking}
During training of encoder-decoder models, the target tokens in the decoder are causally masked to prevent them from looking at future information. 
The causal masking can be represented as a mask matrix $M$:
\[
M_{ij} = 
\begin{cases} 
0
& \text{if } 
j \leq i
 \\
-\infty
& \text{otherwise} 
\end{cases}
\]
\noindent that is summed with the attention matrix before the softmax operator to make sure that each element $i$ can only attend to itself and elements before it (i.e., $j \leq i$), obtaining
\begin{equation*}
    A(Q,K,V) = softmax \left( \frac{QK^T}{\sqrt{d_k}} + M \right) V
\end{equation*}

In standard settings, causal masking is also applied in the DFP models where both previous tokens $\mathbf{Y}$ and the input audio representation $\mathbf{X}$ are masked. Therefore, the decoder self-attentions implement the above masking strategy on the concatenated sequence $\mathbf{X} \mathbin\Vert \mathbf{Y}_{0, ..., i-1}$ (Figure \ref{fig:decoder-only} and \ref{fig:decoder-prepend}). Recent works \citep{10389705} propose an alternative solution for causal masking, where only the previous tokens are masked while each element of the speech sequence can attend to each other. In this case, the causal mask $M$ becomes:

\[
M_{ij} = 
\begin{cases} 
0
& \text{if } j 
\leq
i \text{ or } j < N \\
-\infty
& \text{otherwise} 
\end{cases}
\]

\noindent where $N$ is the length of the speech sequence $\mathbf{X}$. This enables speech tokens to attend to all other speech tokens, including subsequent ones, in the decoder self-attention layers, as it happens in the self-attention of the speech encoders in encoder-decoder models.

\section{Experimental Settings}
\label{sec:exp-sett}
\subsection{Data}
The MuST-C data set is derived from TED talks with English audios transcribed and translated into 8 languages. We trained ASR models using its English transcripts while for ST we also used all 8 target languages, namely Dutch (nl), French (fr), German (de), Italian (it), Portuguese (pt), Romanian (ro), Russian (ru) and Spanish (es). More specifically, we trained two bilingual ST models translating English speech into Spanish and German texts respectively, and a single multilingual ST model translating into all 8 target languages. 

One limitation of MuST-C is that its speech data is English only. In order to compare the models on non-English speech, we run further experiments on the x-en language directions of the CoVoST2 data. There are 21 non-English languages, e.g., Catalan (ca) and Chinese (zh) for the speech inputs. A complete list of supported languages can be found in \citet{wang2021covost}. For each model architecture, we trained a single multilingual ASR\footnote{English ASR data was excluded from training.} model and a multilingual ST model with transcripts of the 21 non-English languages and the English translations as target inputs, respectively. 

\paragraph{Audio processing.} 
We extract log Mel-filterbank features of size 80 computed every 10ms with a window size of 25ms. The resulting spectrograms are normalized using Utterance-level Cepstral Mean and Variance Normalization (CMVN). During training, we also apply SpecAugment \cite{park19e_interspeech} with frequency and temporal masks of size 27 and 100 respectively.

\paragraph{Text processing.} 
The text is tokenized with unigram models trained using SentencePiece \cite{kudo-richardson-2018-sentencepiece}. On CoVoST2, the vocabulary size for ASR and ST is 32k and 5k respectively. On MuST-C, the vocabulary size is 5k for ASR, 8k for bilingual ST, and 32k for multilingual ST. 

\begin{table*}[th]
\centering
\footnotesize
\addtolength{\tabcolsep}{-2.5pt}
\begin{tabular}{l|l|c|c|c|c|ccc}
\toprule
\multirow{3}{*}{\textbf{Line}} & \multirow{3}{*}{\textbf{Model}} & \multirow{3}{*}{\textbf{\#Params (M)}} & \multicolumn{2}{c|}{\textbf{CoVoST2}} & \multicolumn{4}{c}{\textbf{MuST-C}} \\
\cline{4-9}
& & & \textit{ASR -- WER} ($\downarrow$) & \textit{ST -- BLEU} ($\uparrow$) & \textit{ASR -- WER} ($\downarrow$) & \multicolumn{3}{c}{\textit{ST -- BLEU} ($\uparrow$)} \\ 
& & & ca/de/es/fr 
& ca/de/es/fr-en 
& en 
& en-es & en-de & en-x 
\\ 
\midrule 
1 & cross-attention TF & 71.2 - 98.8 & 23.7 & 25.6 & 12.1 & 26.9 & 22.0 & 25.1 \\
2 & decoder-prepend TF & \multirow{2}{*}{64.9 - 92.5}   & 24.7$^{\dagger4}$ & 24.6$^{\dagger4}$ & 12.4 & 26.9 & 21.1$^{\dagger}$ & 24.6$^{\dagger4}$ \\ 
3 & decoder-only 18L &  & 26.1$^{\dagger4}$ & 24.6$^{\dagger4}$ & 13.2$^{\dagger}$ & 27.4$^{\dagger}$ & 21.9 & 25.3$^{\dagger1}$ \\
\midrule
4 & cross-attention CF-CTC & \multirow{2}{*}{111 - 153} & 19.6 & 29.7 & 10.4 & 29.9 & 25.2 & 28.6 \\
4.1 & \phantom{Th} (+) compr &  & 21.8$^{\ddagger4}$ & 28.5$^{\ddagger4}$ & 10.3 & 30.2 & 25.5 & 28.1$^{\ddagger5}$ \\
5 & decoder-prepend CF-CTC & \multirow{2}{*}{105 - 147} & 19.9 & 29.7 & 10.3 & 30.2 & 25.4 & 28.3$^{\ddagger2}$ \\
5.1 & \phantom{Th} (+) compr &  & 19.9 & 29.7 & 10.4 & 30.7$^{\ddagger}$ & 25.9$^{\ddagger}$ & 28.0$^{\ddagger5}$ \\
6 & decoder-only 32L & 109 - 137 & 24.3$^{\ddagger4}$ & 25.3$^{\ddagger4}$ & 13.2$^{\ddagger}$ & 27.2$^{\ddagger}$ & 22.2$^{\ddagger}$ & 26.8$^{\ddagger8}$ \\
\bottomrule
\end{tabular}
\caption{Comparison between cross-attention, decoder-only and decoder-prepend using Transformer (TF) and Conformer (CF) encoders. CTC compression is denoted by "compr". For multilingual models, we report the average over their target languages, i.e., top-4 high resourced pairs for CoVoST2 and the 8 target languages for MuST-C (the "en-x" column). We evaluate ASR and ST with WER ($\downarrow$) and BLEU ($\uparrow$), respectively. ${\dagger}$(N) and ${\ddagger}$(N) refer to the number (N) of language pairs that are significantly different with $p$ < 0.05 to line (1) and line (4), respectively.}
\label{tab:merge_others}
\end{table*}

\subsection{Model architecture}
For our experiments, we use both Transformer and Conformer \citep{gulati20_interspeech} architectures, the latter being an improved version of Transformer for speech achieving state-of-the-art results. All models in our experiments have 18 layers that are distributed between the encoder-decoder (12 layers for the encoder and 6 layers for the decoder) or the decoder, except in \texttt{decoder-only32L} which contains 32 layers to match its number of parameters to the Conformer models. The embedding layer and the feed-forwad layers have dimensions of 512 and 2048, respectively across all the models, and the number of attention heads is set to 8. Encoder dropout is set as 0.1 for feed-forward, attention, and convolution layers. Also, in the convolution layer of the Conformer, the kernel size of the point- and depth-wise convolutions is set to 31. The smallest model has about 64.9M parameters, whereas the largest one has 153M parameters.

In the experiments with CTC loss, a linear layer having the vocabulary size is added after the 8\textsuperscript{th} encoder layer. The softmax function is applied to this layer and then the CTC loss is computed with a weight of 0.5. When CTC compression is applied, vectors having the same predictions are merged by averaging them, following \citet{gaido-etal-2021-ctc}.

We use Fairseq \cite{ott-etal-2019-fairseq,wang-etal-2020-fairseq} for all the experiments. When using the Conformer encoder, we adopt the implementation from \citet{papi-etal-2024-good} which fixes the padding bugs in the convolution layers and the relative positional encoding.

\subsection{Training and Evaluation}\label{subsec:traineval}
\paragraph{Training.} In all experimental settings, we use the Adam optimizer for training. The learning rate follows a Noam scheduler with a maximum value of $2\times 10^{-3}$ and a linear warmup of 25k steps, after which it follows an inverse square root decay.

On MuST-C, all models are trained with a total batch size of 320k audio frames for at most 100k steps along with an early stopping strategy with patience of 10. For the multilingual ST models, we further prepend a language tag to the translations corresponding to the target language. On CoVoST2, all (multilingual) ASR and ST models are trained with a total batch size of 256k frames for 60k steps. On both datasets, all ST encoders are initialized by the corresponding ASR encoders. In the case of decoder-only, we use the corresponding decoder-only ASR model for initialization, but both the embedding and output layer are randomly initialized. Experiments are run on 4 Nvidia A100-40GB GPUs for about 2 days. The final model is obtained by averaging the last 5 checkpoints.

\paragraph{Evaluation.} We use beam search for inference with beam size of 5 and no-repeat-ngram-size of 5, performed on one Nvidia A100-40GB GPU.\footnote{For efficiency, we used a batch size of 80k frames, but results do not depend on inference batch size \citep{papi-etal-2024-good}.}

ASR models are evaluated by computing word error rate (WER), whereas we use sacreBLEU\footnote{
\#1|c:mixed|e:no|tok:13a|s:exp|v:2.4.2} \cite{post-2018-call} to compute BLEU scores for the ST models. We provide statistical significance tests to major comparisons using bootstrap resampling \citep{koehn-2004-statistical} for ASR and approximate randomization \cite{riezler-maxwell-2005-pitfalls} for ST.

\section{Results}
In this section, we conduct a series of experiments to examine cross-attention and DFP from a variety of angles under comparable data and model size conditions. To begin with, we first discuss the ASR and ST results between cross-attention, decoder-only, and decoder-prepend using transformer and conformer architectures. Then, we present their results under the effect of speech sequence compression (via CTC) and seqKD. In addition, we analyze their difference in terms of generation speed and GPU memory footprint. Finally, we present an ablation study about the causality masking of decoder-only and decoder-prepend.

\subsection{Cross-attention, decoder-only and decoder-prepend}
We present the results in Table \ref{tab:merge_others}. In addition, we compute $p$-values between each configuration against the cross-attention baseline of similar model size. 

\paragraph{Transformer encoder (Lines 1-3).} Compared to both DFP methods, cross-attention on average has stronger ASR and ST results. On the CoVoST2 dataset, its improvement in multilingual ASR and ST reaches 2.4 WER (line 1 vs. line 3) and 1 BLEU point, respectively. On the MuST-C dataset, it is still better than decoder-prepend (line 2) and all settings of decoder-only (line 3), except ST on the en-es and en-x directions. These differences are significant for at least one language pair with $p$ < 0.05. Despite its slightly stronger ST results, decoder-only falls behind decoder-prepend in ASR (line 3 vs. line 2), whose WER is 1.4 and 0.8 points lower on CoVoST2 and MuST-C, respectively. The mixed results on MuST-C, especially on ST, indicate the importance of having several test sets, modeling choices and tasks for evaluation.

\paragraph{Conformer with auxiliary CTC (Lines 4-6).}
Since the Conformer model outperforms Transformer in speech processing tasks \citep{gaido-etal-2022-efficient}, we conducted experiments also leveraging this architecture.
Additionally, we apply auxiliary CTC loss on the transcripts during training (see Section \ref{sec:exp-sett} for more details). 

In Table \ref{tab:merge_others}, we can observe that both cross-attention (line 4) and decoder-prepend (line 5) have similar ASR and ST results. On CoVoST2, both models have the same 29.7 BLEU points in multilingual ST, whereas cross-attention has a little advantage of 0.3 WER in ASR. On MuST-C ASR, on the contrary, decoder-prepend has a WER of 0.1 point lower. In terms of ST, decoder-prepend is up to 0.3 BLEU points higher, but it is also 0.3 BLEU points lower in the multilingual case. None of these differences are statistically significant with $p$ < 0.05.
Regarding decoder-only, we scale up its number of layers from 18 to 32 to match the Conformer size. Such scaling improves its overall performance substantially, resulting in a maximum improvement of 1.8 WER in ASR and 0.7 BLEU points in ST (line 3 vs line 6). Despite the improvements, the decoder-only configuration is about 2 points worse than the others in all evaluation settings. 

The above results show that decoder-prepend is on par with cross-attention but not better. This extends to the experiment applying the auxiliary CTC loss to the Conformer encoder. Furthermore, our results clearly show that a properly designed speech encoder, such as the Conformer, substantially improves the performance over a plain decoder-only model of similar size for both ASR and ST. Because of its competitive performance, we further compare decoder-prepend with cross-attention in the following sections.

\begin{figure*}[th]
    \centering
    \input{seqkd_barplot}
    \caption{Comparison between cross-attention and decoder-prepend using sequence-level knowledge distillation (seqKD) for ST on MuST-C en-de. "\textit{TF}", "\textit{CF-CTC}" and "\textit{CF-compr}" refer to Transformer, Conformer with auxiliary CTC loss and Conformer with CTC compression, respectively.}
    \label{fig:seqKD_merge}
\end{figure*}
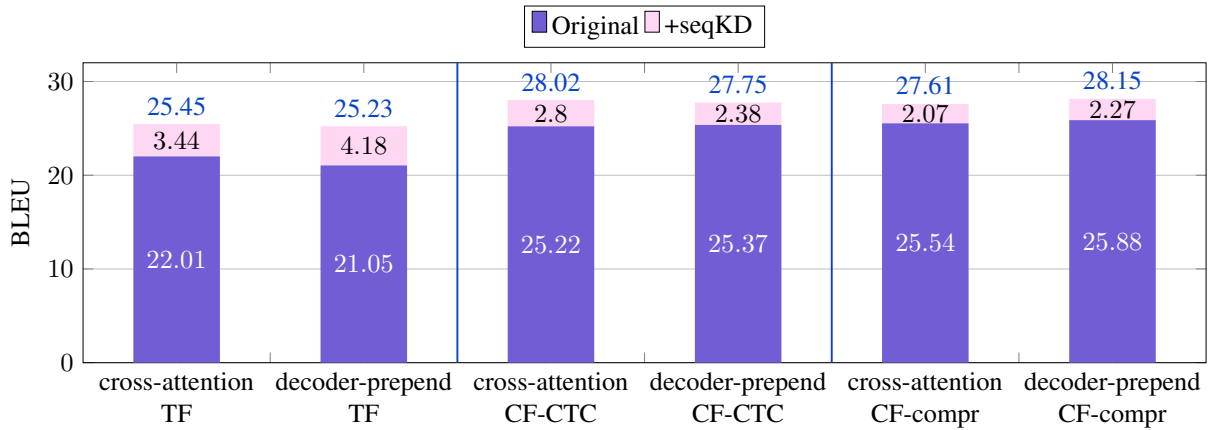

\subsection{Effect of audio sequence compression}
Despite the similar quality achieved when auxiliary CTC loss is applied during training (lines 4-5), the behaviour of cross-attention and decoder-prepend significantly differ when CTC compression is also applied. On the CoVoST2 dataset, compression on cross-attention (line 4.1) causes a degradation of 2.2 points in WER and 1.2 points in BLEU (with $p$ < 0.05), whereas it does not cause harm to decoder-prepend (line 5.1). On the MuST-C dataset, both cross-attention and decoder-prepend get better in bilingual ST and worse in multilingual ST after applying compression while remaining stable for ASR. Despite the similar pattern, the improvements of decoder-prepend are slightly bigger while its degradations are smaller. Thus, overall, decoder-prepend better leverages CTC compression.

Our results indicate that applying CTC compression to decoder-prepend is more beneficial than to cross-attention but it is not sufficient to claim that decoder-prepend is better. Cross-attention with CTC (line 4) and decoder-prepend with compression (line 5.1) have almost identical results on CoVoST2, whereas, on MuST-C, cross-attention is 0.6 BLEU points better in multilingual ST despite being 0.8 BLEU points worse in bilingual ST.

\subsection{Effect of seqKD}\label{subsec:seqKD}
SeqKD helps to reduce the translation data complexity by making the target sentence more monotonically aligned to its source sentence \cite{zhou2019understanding}. This makes seqKD not only useful for improving non-autoregressive translation but also end-to-end ST \cite{inaguma-etal-2021-source}, which has an additional challenge brought by the modality gap. Despite its usefulness, there is a lack of studies about applying seqKD on ST models using DFP. In the following, we fill the gap with experiments on the MuST-C en-de language direction, where cross-lingual alignment is more complicated, to demonstrate better the effect of the seqKD data. We employ NLLB 3.3B \cite{costa2022no} to machine translate the English transcripts in the training set into German as our seqKD data.

Figure \ref{fig:seqKD_merge} presents the results on cross-attention and decoder-prepend\footnote{We exclude decoder-only from these experiments since decoder-prepend has been shown better ASR and ST qualities than decoder-only.}. For each model configuration, we train once using the original data, denoted by "Original", and once using the combined data, denoted by \textit{"Original+seqKD"}. As we can observe, training on the combined data brings a substantial gain of more than 2 BLEU points ("+seqKD") in all configurations. This indicates that seqKD also works effectively on decoder-prepend. 

Despite the remarkable gain, there are no clear indications of whether cross-attention or decoder-prepend benefits more from seqKD data. In fact, in the case of Transformer, decoder-prepend gets a higher BLEU improvement (4.18 vs 3.44) with respect to cross-attention. However, in the case of Conformer, we observe a different behaviour: the gain for decoder-prepend is higher only when CTC compression is applied. In terms of their best BLEU scores, both cross-attention and decoder-prepend are very similar: cross-attention using Transformer is 0.22 BLEU points higher than decoder-prepend, whereas the conformer results show the opposite: 
decoder-prepend with compression is 0.13 BLEU points better than cross-attention (28.15 vs 28.02).

\subsection{Generation speed and memory footprint}
In addition to ASR and ST performances, we evaluate our models in terms of generation speed and GPU memory footprint. For each metric, we compute the average over the 8 language pairs using the multilingual ST on MuST-C, which has longer inputs and covers diverse languages, providing more robust statistics. We report the relative value using cross-attention with Transformer encoder as the baseline in Table \ref{tab:speed_memory}. The batch size and GPU settings follow those in Section \ref{subsec:traineval}, except that only one GPU is used. 

\begin{table}[ht]
\centering
\resizebox{0.48\textwidth}{!}{
\begin{tabular}{l|c|c|c}
\toprule
\multirow{2}{*}{\textbf{Model}} & \multirow{2}{*}{\textbf{\#Params}} & \multicolumn{2}{c}{\textbf{Ratio}} \\
\cline{3-4}
& & \textit{speed} $\uparrow$ & \textit{memory} $\downarrow$ \\
\midrule
cross-attention TF & 98.8M & 1 & 1 \\
decoder-prepend TF & \multirow{2}{*}{92.5M} & 0.96 & 1.59 \\
decoder-only 18L &  & 0.83 & 3.01 \\
\midrule
cross-attention CF-CTC & \multirow{2}{*}{139M} & 0.97 & 2.16 \\ 
\phantom{Th} (+) compr &  & 0.98 & 1.88 \\
decoder-prepend CF-CTC & \multirow{2}{*}{133M} & 0.94 & 3.11 \\ 
\phantom{Th} (+) compr &  & 0.96 & 2.50 \\
decoder-only 32L & 136M & 0.70 & 5 \\ 
\bottomrule
\end{tabular}
}
\caption{A comparison between cross-attention and DFP in terms of model parameters, relative generation speed (tokens/s) and relative GPU memory footprint. Other acronyms follow Table \ref{tab:merge_others}.}
\label{tab:speed_memory}
\end{table}

Compared to cross-attention, decoder-prepend has fewer parameters, i.e., 98.8M vs 92.5M, but is 4\% slower. If we allocate all encoder parameters to the decoder, i.e., \texttt{decoder-only (18L)}, the resulting model is even 17\% slower than cross-attention while requiring about three times as much GPU memory. Similar patterns could be found in conformer with CTC, where cross-attention is slightly faster and less memory demanding than its decoder-prepend despite having 6M more parameters. Again, decoder-only, i.e., \texttt{decoder-only (32L)}, appears to be the worst. It has only 70\% generation speed and requires 5 times as much GPU memory footprint to the baseline. It is worth noting that \texttt{cross-attention CF-CTC} still remains faster and more memory efficient than \texttt{decoder-only (18L)} despite having 43\% more parameters. As expected, CTC compression (\texttt{compr}) makes the generation faster and reduces the GPU memory footprint. The improvement to cross-attention and decoder-prepend in speed is about 1\% and 2\%, respectively, whereas it is respectively about 13\% and 19\% in memory footprint. Decoder-prepend has a bigger improvement, but its overall performance is still behind that of cross-attention. 

Despite the removal of the cross-attention layers, our study reveals that DFP is still worse in terms of generation speed and memory footprint. The quadratic time and memory complexity of self-attention in sequence length is a more severe issue for DFP when considering speech inputs.

\subsection{(Audio) Causality masking in decoder-only and decoder-prepend}
\begin{table*}[tb]
\centering
\resizebox{\textwidth}{!}{
\begin{tabular}{l|c|c|c|c|ccc}
\toprule
\multirow{3}{*}{\textbf{Model}} & \multirow{3}{*}{\textbf{\#Parameters}} & \multicolumn{2}{c|}{\textbf{CoVoST2}} & \multicolumn{4}{c}{\textbf{MuST-C}} \\
\cline{3-8}
& & \textit{ASR -- WER} ($\downarrow$) & \textit{ST -- BLEU} ($\uparrow$) & \textit{ASR -- WER} ($\downarrow$) & \multicolumn{3}{c}{\textit{ST -- BLEU} ($\uparrow$)} \\ 
& & ca/de/es/fr & ca/de/es/fr-en & en & en-es & en-de & en-x \\ 
\midrule 
decoder-prepend TF & \multirow{4}{*}{64.9M - 92.5M} & 24.7 & 24.6 & 12.4 & 26.9 & 21.1 & 24.6 \\
\phantom{Th} (-) causal mask &   & 24.6 & 24.9 & 13.2$^{\dagger}$ & 25.8$^{\dagger}$ & 20.8 & 16.5$^{\dagger8}$\\ 
decoder-only 18L &  & 26.1 & 24.6 & 13.2 & 27.4 & 21.9 & 25.3 \\
\phantom{Th} (+) causal mask &  & 28.7$^{\dagger4}$ & 22.2$^{\dagger4}$ & 13.9$^{\dagger}$ & 26.0$^{\dagger}$ & 20.1$^{\dagger}$ & 24.2$^{\dagger8}$ \\
\midrule
decoder-prepend CF-CTC & \multirow{4}{*}{105M - 147M} & 19.9 & 29.7 & 10.3 & 30.2 & 25.4 & 28.3  \\
\phantom{Th} (-) causal mask &  & 20.1 & 29.9 & 10.6 & 30.2 & 25.3 & 28.3\\
\phantom{Th} (+) compr &  & 19.9 & 29.7 & 10.4 & 30.7 & 25.9 & 28.0 \\
\phantom{Th}\phantom{Th} (-) causal mask &  & 19.9 & 29.7 & 10.4 & 30.7 & 25.5 & 28.1\\
decoder-only 32L & \multirow{2}{*}{109M - 137M} & 24.3 & 25.3 & 13.2 & 27.2 & 22.2 & 26.8\\
\phantom{Th} (+) causal mask &  & 26.6$^{\dagger4}$ & 23.3$^{\dagger4}$ & 14.4$^{\dagger}$ & 26.0$^{\dagger}$ & 20.2$^{\dagger}$ & 25.5$^{\dagger8}$ \\ 
\bottomrule
\end{tabular}
}
\caption{Causality masking in decoder-only and decoder-prepend. ${\dagger}$(N) refers to the number (N) of language pairs that are significantly different with $p$ < 0.05 to its baseline. Other acronyms follow Table \ref{tab:merge_others}.}
\label{tab:causalmask}
\end{table*}

In previous sections, we presented each DFP configuration using its optimal causal masking strategy: 1) causal masking is not applied on decoder-only, whereas 2) it is applied on decoder-prepend. In the following, we provide an ablation study of causal masking, which is summarised in Table \ref{tab:causalmask}. The significance tests are computed between the pairs with and without causal masking.

As we can observe, decoder-only (both \texttt{18L} and \texttt{32L}) performs worse on all experimental settings when causal masking is applied. On CoVoST2, the performance degrades by at least 2 points, whereas the degradation can be up to 2 
points on MuST-C. This indicates the importance of allowing the speech frames to attend each other in decoder-only models. Our finding is in accordance with the conclusion drawn by \citet{gupta24_interspeech} for ASR, and we further extend it for ST. 

When causal masking is removed from decoder-prepend (\texttt{decoder-prepend TF}), we observe a performance degradation of 1.2 WER and up to 0.9 BLEU points on MuST-C ASR and bilingual ST, respectively. What is even worse is the degradation of 8.1\footnote{The degradation is similar when the experiment is repeated with another random seed.} BLEU points in multilingual ST. On the CoVoST2 dataset, however, removal of causal masking causes little improvement to both ASR and ST. In the case of Conformer, there are almost no performance changes on the CoVoST2 dataset when causal masking is removed, but a small degradation of 0.4 BLEU points (25.9 $\rightarrow$ 25.5) on the MuST-C en-de direction when CTC compression is also applied. 

Therefore, our results lead to two interesting observations. Firstly, the behaviour of DFP models are quite different with causal masking, depending on whether a speech encoder is used or not. We hypothesise that the non-adversarial effect of causal masking on decoder-prepend is attributed to the self-attention within the speech encoder, which allows full attention within the speech frames. Secondly, applying causal masking on decoder-prepend is likely to help improving model performance on longer speech inputs, such as the MuST-C dataset.

\section{Conclusion}
 In this paper, we aim to validate the modeling choice of using DFP (decoder-only and decoder-prepend) over cross-attention to integrate speech into decoder-only LLMs for S2T tasks. In order to perform a controlled comparison under limited computational budget, we train all models from scratch without using large pretrained models. Our series of comparisons, including mono/bi/multi-lingual settings, indicate that DFP does not consistently outperform cross-attention in ASR and ST quality, and that cross-attention is
 more efficient in terms of generation speed and GPU memory footprint. Our studies further suggest that: (1) decoder-prepend with a strong speech encoder is more efficient than decoder-only of similar size, and (2) a variety of test sets, language pairs (and directions) as well as tasks, e.g., bi/multi-lingual ST models, are needed to validate the effective scope of a S2T technique, such as causal masking.  

\paragraph{Future Work.} In addition to scaling up the data and model sizes, we leave the comparisons of several interesting aspects of S2T models in future works. These include 1) zero-shot transfer \cite{tsiamas-etal-2024-pushing}, 2) performance under segmental inputs and augmentations \cite{tsiamas22_interspeech,lam-etal-2022-sample,lam2023make}, 3) simultaneous setting \cite{ahmad-etal-2024-findings,papi2024real} and 4) additional tasks such as spoken language understanding \cite{lee2024speech} and spoken question answering \cite{you-etal-2022-end,zufle2024contrastive}.

\section{Limitations}
We note the limitations of our experiments. Firstly, our study is based on a single scale point, i.e., without covering a wide range of model parameters, so that the conclusion might change with scale. Secondly, LLMs have slightly different modeling options than our settings, such as having instructions between the speech inputs and the target texts as well as using rotary positional encoding rather than absolute. Given the limited computational budget, we could not include additional comparisons but our studies have confirmed existing findings and brought it to wider scopes compared to previous works.

\section*{Acknowledgments}
Marco Gaido's work has been funded by the PNRR project FAIR -  Future AI Research (PE00000013),  under the NRRP MUR program funded by the NextGenerationEU. Sara Papi and Luisa Bentivogli's work has been supported by the European Union's Horizon research and innovation programme under grant agreement No 101135798, project Meetween (My Personal AI Mediator for Virtual MEETtings BetWEEN People). 
Barry Haddow and Tsz Kin Lam's work were funded by UK Research and Innovation (UKRI) under the UK government’s Horizon Europe funding guarantee (grant number 10039436: UTTER). 

For the computational resources used in this research we acknowledge the CINECA award (MAGIS) under the ISCRA initiative and the Baskerville\footnote{https://www.baskerville.ac.uk/} Tier 2 HPC service. Baskerville was funded by the EPSRC and UKRI through the World Class Labs scheme (EP/T022221/1) and the Digital Research Infrastructure programme (EP/W032244/1) and is operated by Advanced Research Computing at the University of Birmingham.
%

\bibliography{anthology,custom}

\end{document}

%% file: seqkd_barplot.tex
    \begin{tikzpicture}[scale=0.9]
    \begin{axis}[
        width=18cm,
        height=6cm,
        ybar stacked,
        ylabel=BLEU,
        ymin=0,
        ymax=32,
        xtick={1.5,3.5,5.5},
        xticklabels={},
        extra x ticks={1,2,3,4,5,6},
        extra x tick labels={
            cross-attention\\TF,
            decoder-prepend\\TF,
            cross-attention\\CF-CTC,
            decoder-prepend\\CF-CTC,
            cross-attention\\CF-compr,
            decoder-prepend\\CF-compr
        },
        legend style={at={(0.5,1.05)}, anchor=south, legend columns=-1},
        ymajorgrids=true,
        xmajorgrids=false,
        nodes near coords,
        every node near coord/.append,
        bar width=36pt,
        x tick label style={align=center},
        enlarge x limits=0.1,
    ]
    \definecolor{customlightblue}{HTML}{BFE4FE}
    \definecolor{custompink}{HTML}{FED7F3}
    \definecolor{customlightpurple}{RGB}{204,204,255}
    \definecolor{custompurple}{HTML}{725DD4}
    \definecolor{customblue}{HTML}{0944C8}
    
    \draw[customblue, thick] (axis cs:2.5,0) -- (axis cs:2.5,32);
    \draw[customblue, thick] (axis cs:4.5,0) -- (axis cs:4.5,32);
    
    \addplot[fill=custompurple, draw=none, nodes near coords={\color{white}\pgfmathprintnumber\pgfplotspointmeta}] coordinates {
        (1,22.01) (2,21.05) (3,25.22) (4,25.37) (5,25.54) (6,25.88)
    };
    
    \addplot[draw=none, fill=custompink] coordinates {
        (1,3.44) (2,4.18) (3,2.8) (4,2.38) (5,2.07) (6,2.27)
    };
    
    \node[above, text=customblue] at (axis cs:1,25.45) {25.45};
    \node[above, text=customblue] at (axis cs:2,25.23) {25.23};
    \node[above, text=customblue] at (axis cs:3,28.02) {28.02};
    \node[above, text=customblue] at (axis cs:4,27.75) {27.75};
    \node[above, text=customblue] at (axis cs:5,27.61) {27.61};
    \node[above, text=customblue] at (axis cs:6,28.15) {28.15};
    
    \legend{Original,+seqKD}
    
\end{axis}
\end{tikzpicture}